\documentclass[twoside,11pt]{article}

\usepackage{jmlr2e}
\usepackage{graphicx}
\usepackage{amsmath}
\usepackage{wrapfig}
\usepackage[utf8]{inputenc} 
\usepackage[T1]{fontenc}    
\usepackage{hyperref}       
\usepackage{url}            
\usepackage{booktabs}       
\usepackage{amsfonts}       
\usepackage{nicefrac}       
\usepackage{microtype}      

\ShortHeadings{Scalable Modeling of Multivariate Longitudinal Data}{}
\firstpageno{1}

\begin{document}

\title{Scalable Modeling of Multivariate Longitudinal Data for Prediction of Chronic Kidney Disease Progression}

\author{\name Joseph Futoma  \email jdf38@duke.edu \\
       \addr Dept. of Statistical Science \\
       Duke University \\
       Durham, NC 27708
       \AND
	\name Mark Sendak  \email mark.sendak@dm.duke.edu \\
       \addr Institute for Health Innovation \\
       School of Medicine \\
       Duke University \\
       Durham, NC 27708 
       \AND
	\name C. Blake Cameron  \email blake.cameron@dm.duke.edu \\
       \addr Division of Nephrology \\
       Duke University \\
       Durham, NC 27708 
       \AND
	\name Katherine Heller  \email kheller@stat.duke.edu \\
       \addr Dept. of Statistical Science \\
       Duke University \\
       Durham, NC 27708 
       } 

\maketitle

\begin{abstract}
Prediction of the future trajectory of a disease is an important challenge for personalized medicine and population health management.  However, many complex chronic diseases exhibit large degrees of heterogeneity, and furthermore there is not always a single readily available biomarker to quantify disease severity.  Even when such a clinical variable exists, there are often additional related biomarkers routinely measured for patients that may better inform the predictions of their future disease state.  To this end, we propose a novel probabilistic generative model for multivariate longitudinal data that captures dependencies between multivariate trajectories.  We use a Gaussian process based regression model for each individual trajectory, and build off ideas from latent class models to induce dependence between their mean functions.  We fit our method using a scalable variational inference algorithm to a large dataset of longitudinal electronic patient health records, and find that it improves dynamic predictions compared to a recent state of the art method.  Our local accountable care organization then uses the model predictions during chart reviews of high risk patients with chronic kidney disease.
\end{abstract}

\section{Introduction}
The ability of healthcare organizations to make accurate predictions about individuals' future health is becomingly increasingly important with the adoption of accountable care and alternative payment models (\cite{CMS}).  In particular, management of complex, chronic diseases such as cardiovascular disease, diabetes, and chronic kidney disease is especially difficult, because selection of optimal therapy may require integration of multiple conditions and risk factors that are each considered in isolation under current approaches to care.  Additionally, individuals with multiple chronic conditions are among both the most expensive and highest utilizers of healthcare services (\cite{Johnson}).

Accountable care organizations (ACOs) are institutions that aim to link provider reimbursement with quality metrics and reductions in total cost of healthcare services to a specific population of patients.  These organizations need personalized prediction tools capable of flagging specific patients in their populations at the highest risk of having poor outcomes (\cite{Parikh}).  Much of the data required to develop such tools are already being routinely collected, due to the recent widespread adoption of Electronic Health Records (EHRs).  However, in order to be clinically relevant the prediction methods will need to be flexible enough to accommodate the inherent limitations of operational EHR data (\cite{Hersh}), dynamically update predictions as more information is collected, and scale seamlessly to the massive size of modern health records. 

Prediction of future disease trajectory is a challenging problem.  One difficulty is the many underlying sources of variability that can drive the different potential manifestations of the disease.  For instance, the underlying biological mechanisms of the disease can give rise to latent disease subtypes, or groups of individuals with shared characteristics (\cite{subtype}).  For most complicated diseases, there are no clear definitions of subtypes, so this must be inferred from the data.  In addition, there are individual-specific sources of variability that may not be directly observed, such as behavioral factors, environmental conditions, or temporary infections.  Another challenge is the fact that observations are irregularly sampled, asynchronous, and episodic, precluding the use of many time series methods developed for data regularly sampled at discrete time intervals.  The large degree of missing data, especially when modeling multivariate longitudinal data, also presents complications.  The task is made even more difficult when the primary data source is the EHR rather than a curated registry, as even defining a relevant cohort of patients to model can take extensive review by a clinical expert.  

\begin{figure*}
  \includegraphics[width=1.0\textwidth,height=7cm]{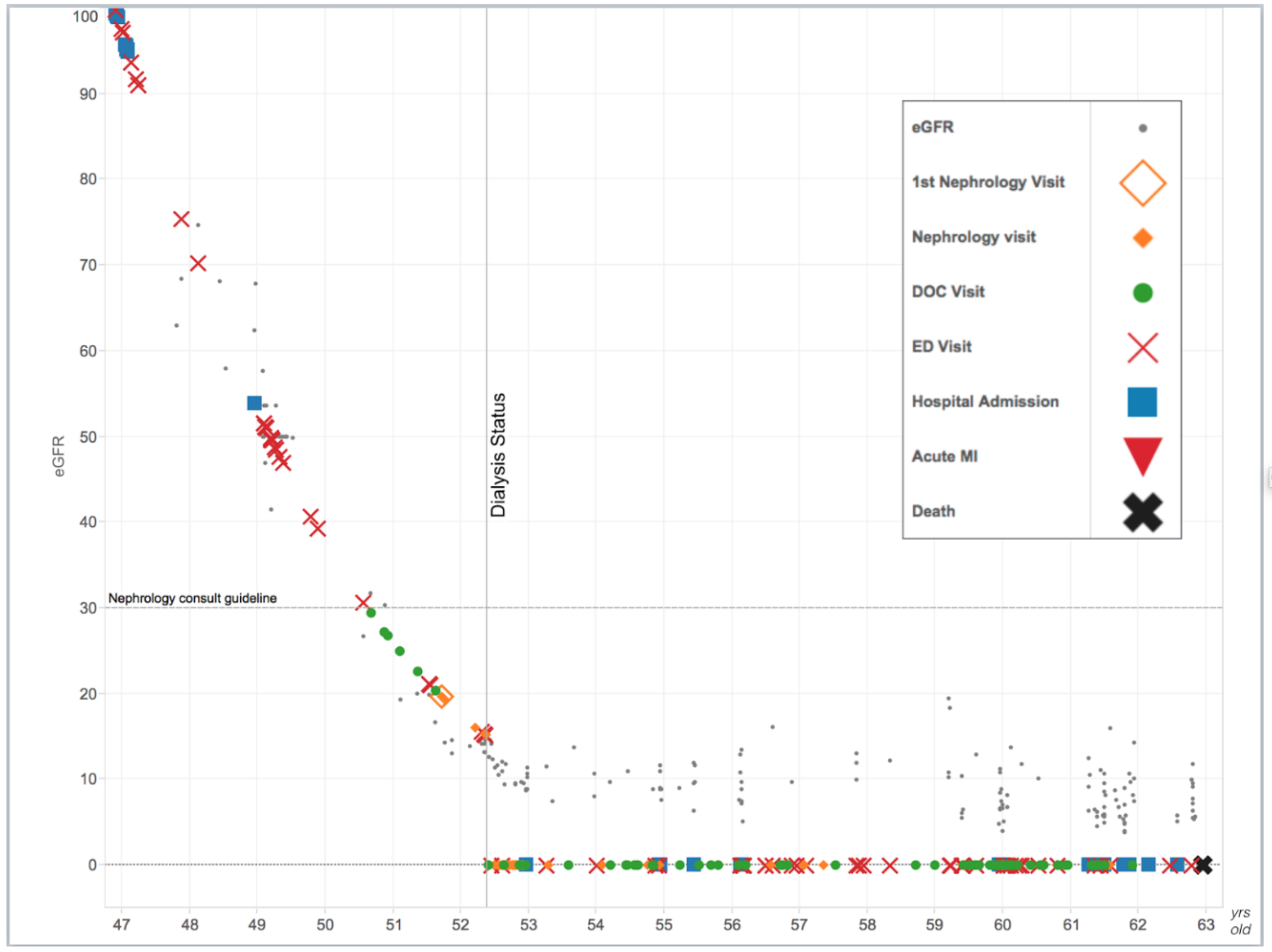}
  \caption{Clinical course of a patient who experienced a rapid progression of CKD as well as other serious health events.  Y-axis denotes estimated glomerular filtration rate (eGFR), an estimate of overall kidney function (60-100 is normal, $<$60 indicates clinically significant kidney disease).  X-axis denotes patient age, in years.  Markers denote adverse events and health system use.}
\end{figure*}

We collaborated with Duke Connected Care, the ACO affiliated with the Duke University Health System, to create models to predict disease progression in chronic kidney disease (CKD).  CKD is characterized by a slow and typically symptomless loss of kidney function over time, resulting in complications that can cause poor health, premature death, increased health service utilization, and excess economic costs.  A person's estimated glomerular filtration rate (eGFR) is an approximation of overall kidney function and is calculated using a common clinical laboratory test (serum creatinine or cystatin C) and demographic information (age, sex and race) (\cite{eGFR}).  Impairment of a person's eGRF is used to define and stage CKD.  In addition, a number of other routinely measured laboratory values may be used to help detect abnormal or declining kidney function.  

Providing optimal care for CKD patients is a challenging problem for healthcare providers. Even though CKD can be easily identified using simple, eGFR-based laboratory criteria, it is often not recognized (\cite{Szczech}, \cite{Tuot}).  Predicting which patients with recognized CKD will progress to kidney failure (requiring dialysis or kidney transplantation), or suffer from other complications such as early death from heart attack or stroke, is a difficult task (\cite{Mend}).  Additionally, oftentimes providers fail to prescribe appropriate preventive treatment to slow disease progression or address complications (\cite{Smart}).
 
These traits make CKD an ideal condition to develop disease progression models that can be used in high-impact care management programs. The difficulties with CKD care are best explained with a representative clinical case, illustrated in Figure 1.  A 47 year-old man makes first contact with the health system with normal kidney function, although he possesses several risk factors for developing CKD in the future.  He receives sufficient medical care over the next few years to detect that his kidney function is rapidly deteriorating (normal annual rate of kidney function loss at his age is only about 1-2\%).  Due to other more pronounced medical conditions, his kidney disease goes unnoticed and he does not receive any treatment to slow progression to total kidney failure.  He is finally referred to a kidney specialist at age 52, more than a year after his kidney function has fallen below the recommended referral threshold.  There is now not enough time to make advanced preparations for kidney failure, such as pre-emptive kidney transplantation or at-home dialysis, and kidney failure is inevitable.  Less than 90 days from the first specialist appointment, he requires hospitalization for emergency dialysis initiation, a traumatizing procedure that also makes him among the most expensive class of patient to treat (\cite{Johnson}).  He suffers multiple cardiovascular complications from kidney failure while on dialysis over the next decade, ultimately dying at age 63. This patient story is riddled with missed opportunities that could have been acted upon by a care management program that uses accurate predictions from machine learning models for disease progression.

Our aim in this work is to develop broadly applicable statistical methods that flexibly model multivariate longitudinal data in order to jointly predict future disease trajectories with other related variables, such as patient symptoms or relevant laboratory values.  To accomplish this, we utilize a hierarchical latent variable model that captures dependencies between multivariate longitudinal trajectories.  Our model uses a Gaussian process (GP) with a highly structured mean function to model each longitudinal variable for each individual.  The mean functions for the GPs are made dependent through shared latent variables.  Using our model, we study a large cohort of patients with CKD from the Duke University Health System EHR and make predictions about the future trajectory of their disease severity, as measured by eGFR, along with five other commonly recorded laboratory values that are known to be affected by CKD.  Our inference algorithm scales well to the large dataset, and makes accurate predictions that outperform a state-of-the-art baseline.

\section{Related Work}

Within the medical literature, the vast majority of predictive models are cross-sectional and only consider features at or up until the current time to predict outcomes at a fixed point in the future.  Typically, these models only attempt to explain variability in the outcome of interest by conditioning on baseline covariates.  This precludes the ability to generate dynamic individualized predictions, making them difficult to use for medical decision making in practice.  For instance, \cite{Tangri} is a commonly referenced Cox proportional hazards model for predicting time to kidney failure.  

Within the statistics and machine learning communities, Markov models of many varieties are frequently used to generate dynamic predictions, e.g. autoregressive models, HMMs, and state space models (\cite{Murphy}).  However, these methods are generally only applicable in settings with discrete, regularly-spaced observation times, and in most applications the data consists of a single set of multivariate time series (e.g. financial returns), not a large collection of sparsely sampled multivariate time series (in our setting, disease or lab trajectories).  GPs have been commonly been used in settings with continuous time observations, see e.g. (\cite{GPtime}) for a thorough overview.  Since they are prior distributions over functions, they are a natural modeling choice for disease trajectories, however, accurate forecasts for GPs require careful specification of the mean functions (\cite{Shi}).  

There are several related works that tackle the problem of dynamic predictions with medical applications in mind.  (\cite{Riz}) construct joint models with a focus on updating dynamic predictions about time to death as more values of a longitudinal biomarker are observed.  They also account for individual heterogeneity using random effects.  Related work on joint modeling (\cite{PL}) use a mixture model to address heterogeneity.  Although a few works from this joint modeling literature address multivariate data, typically the only dependency among variables is simplistic, coming from a shared vector of random effects.  Most similar to our work is (\cite{Schulam}), who present a model for a univariate marker of disease trajectory using a GP with highly structured mean function.

There has been much recent interest in machine learning in modeling EHRs and other types of healthcare data.  For instance, (\cite{Lian}) use hierarchical point processes to predict hospital admissions, and (\cite{Raj}) develop a dynamic factor model to learn relationships between diseases and predict future diagnosis codes.  Some related papers use multitask GPs to model multivariate, longitudinal clinical data (\cite{Durichen}, \cite{Ghassemi}).  However, these works use independently trained models for each patient and do not hierarchically share information across patients.  This worked well in their examples where there were a large number of observations for each patient and little missing data, but would not work well with our much sparser EHR data.  Closest to our work in the application is (\cite{Perotte}), who predict time of progression from CKD stage 3 to stage 4.  Although they use a standard Kalman filter to model multivariate laboratory data, this is not the focus of their work and will not be as flexible as our GP-based models.  Finally, \cite{UAI} present a method for jointly modeling CKD disease trajectory with adverse events.

\section{Proposed Multivariate Disease Trajectory Model}

Our proposed hierarchical latent variable model jointly models each patient's multivariate longitudinal data by using a GP for each individual variable, with shared latent variables inducing dependence between the mean functions.  Note that our model reduces to the method presented in (\cite{Schulam}) in the univariate setting. 

Let $\vec{y_i}(t) = (y_{i1}(t),\dots,y_{iP}(t))^\top \in \mathbb{R}^P$ denote the $P-$dimensional trajectory of measurements for individual $i$, and let $\vec{y_{ip}} = \{y_{ip}(t_{ipj})\}_{j=1}^{n_{ip}}$ be the $n_{ip}$ observations for variable $p$ at times $t_{ipj}$ for this individual.  Let $c_i$, $\vec{z_i}$, $\vec{b_i}$, and $\vec{f_i}$ be latent variables specific to individual $i$, to be defined subsequently.  We will assume for each person that the longitudinal variables are conditionally independent given these latent variables: $p(\vec{y_i} | c_i, \vec{z_i}, \vec{b_i}, \vec{f_i} ) = \prod_{p=1}^P p(\vec{y_{ip}} | c_i, \vec{z_i}, \vec{b_i}, \vec{f_i} )$.  

For each longitudinal variable, we propose the following generative model:
\begin{align}
y_{ip}(t) &\sim N( \mu_{ip}(t), \sigma_p^2 ) \\
\mu_{ip}(t) &\sim \mathcal{GP}( \Lambda^{(p)} x_i + \Phi_z(t)^\top \beta^{(p)}_{z_{ip}} + \Phi_l(t)^\top b_{ip}, K_p) \\
K_p(t,t') &= a_p^2 \text{exp}\{-l_p^{-1}|t-t'|\} \\
z_{ip} | c_i &\sim \text{Multinomial}(\Psi_{c_i}^{(p)}) 
\end{align}
with common priors across all $P$ variables:
\begin{align}
 c_i &\sim \text{Multinomial}(\pi_i), \; \pi_{ig} = \frac{e^{w_g^\top x_i}}{\sum_{g'=1}^G e^{w_{g'}^\top x_i}} \\
\vec{b_i} &= (b_{i1},\dots,b_{iP})^\top \sim N(0,\Sigma_b).
\end{align}

The first term in the mean function (2) is a population-level fixed effect that uses observed baseline covariates $x_i \in \mathbb{R}^q$, e.g. gender and race, to determine a population-level intercept for each lab, with $\Lambda^{(p)} \in \mathbb{R}^q$ a coefficient vector.  This can easily be extended to accommodate more flexible fixed effects, e.g. a fixed intercept and slope.

The second term in (2) is a subpopulation component, where it is assumed individual $i$'s trajectory for variable $p$ belongs to a latent subpopulation, denoted $z_{ip} \in \{1,\dots,G_p\}$.  Each subpopulation is associated with a unique trajectory; in particular, $\Phi_z(t) \in \mathbb{R}^{d_z}$ is a fixed B-spline basis expansion of time (for simplicity, assumed to be the same for each variable with eight interior knots evenly spaced in time) with $\beta^{(p)}_g \in \mathbb{R}^{d_z}$ the coefficient vector for subpopulation $g$ and variable $p$.  

The prior for each $z_{ip}$ in (4) depends on the individual's ``global'' cluster $c_i \in \{1,\dots,G\}$.  Drawing on ideas from latent structure analysis for multivariate categorical data (\cite{LSA}), the $c_i$ induce dependence among the $P$ trajectory-specific clusters $z_{ip}$, as they have conditionally independent multinomial priors.  Each of the $G$ columns $\Psi_{g}^{(p)}$ in the matrix $\Psi^{(p)} \in \mathbb{R}^{G_p \times G}$ defines a distribution over the $G_p$ values that $z_{ip}$ can take.   The $c_i$ then has a multinomial logistic regression prior in (5) that depends on the baseline covariates $x_i$, where $\{w_g\}_{g=1}^G$ are regression coefficients with $w_1 \equiv$ 0 for identifiability.  

The third term in (3) is a random effects component, allowing for individual-specific long-term deviations in trajectory that are learned dynamically as more data becomes available.  In practice $\Phi_l(t) \in \mathbb{R}^{d_l}$ is a linear expansion of time with $d_l = 2$, so that $b_{ip} \in \mathbb{R}^{d_l}$ is a random slope and intercept vector for patient $i$.  The overall vector $\vec{b_i}$ has a multivariate normal prior distribution in (6), making the random effects dependent across labs.  

Finally, $K_p(t,t') = a_p^2 \text{exp}\{-l_p^{-1}|t-t'|\}$ is the Ornstein-Uhlenbeck (OU) covariance function for the GP each variable, with parameters $a_p, l_p$.  This kernel is well-suited for modeling transient deviations from the mean function, as it is mean-reverting and has no long-range dependence between deviations.  We can rewrite (2) as $\mu_{ip}(t) = \Lambda^{(p)} x_i + \Phi_z(t)^\top \beta_{z_{ip}} + \Phi_l(t)^\top b_{ip} + f_{ip}(t)$, where $f_{ip}(t) \sim \mathcal{GP}(0,K_p)$, so that we can explicitly represent the short-term deviations $f_{ip}$ from the GP in order to learn them during inference.  

\section{Inference} 

The computational problem associated with fitting our model is estimation of the posterior distribution of latent variables and model parameters given the observed data.  As is common with complex probabilistic generative models, exact computation of the posterior is intractable and requires approximation to compute.  To this end, we develop a mean field variational inference [Jordan et al., 1999] algorithm to find an approximation to the posterior. 

Variational inference converts the task of posterior inference into an optimization problem of finding a distribution $q$ in some approximating family of distributions that is close in KL divergence to the true posterior.  The problem is equivalent to maximizing what is known as the evidence lower bound (ELBO) [Bishop, 2006]: 
\begin{equation}
\mathcal{L}(q) = E_q[\text{log } p(y,z,b,f,c,\Theta) - \text{log } q(z,b,f,c,\Theta)],
\end{equation}
which forms a lower bound on the marginal likelihood $p(y)$ of our model.

\subsubsection*{Variational Approximation}

The model parameters to be learned are $\Theta = \big\{ \{ \Lambda^{(p)},\beta^{(p)},\Psi^{(p)}, a_p, l_p, \sigma_p^2 \}_{p=1}^P, W, \Sigma_b \big\}$, and the local latent variables specific to each person are their global cluster assignment $c_i$, subpopulation assignments $z_{ip}$, random effects $\vec{b_i}$ and structured noise functions $f_{ip}$.  The joint distribution for our model can be expressed as:
\begin{align}
p(y,z,b,f,c,\Theta) &= p(\Theta) \prod_{i=1}^N p(b_i | \Theta) p(c_i | \Theta) \prod_{p=1}^{P} p(\vec{y_{ip}} | z_{ip}, b_{ip}, f_{ip} , \Theta) p(z_{ip} | c_i, \Theta)   p(f_{ip} | \Theta) 
\end{align}
We make the mean field assumption for the variational distribution, which assumes that in the approximate posterior $q$, all the latent variables are independent.  This implies that $q(z,b,f,v,\Theta) = q(\Theta) \prod_{i=1}^N q_i(c_i,z_i,b_i,f_i)$, where:
\begin{equation}
q_i(c_i,z_i,b_i,f_i) = q_i(c_i | \nu_{c_i})  q_i(b_i | \mu_{b_i}, \Sigma_{b_i}) \prod_{p=1}^P q_i(z_{ip} | \nu_{z_{ip}}) q_i(f_{ip} | \mu_{f_{ip}}, \Sigma_{f_{ip}})
\end{equation}
The assumed variational distributions for each latent variable are in the same family as their prior distribution, i.e. $c_i$ and the $z_{ip}$ are multinomials, $b_i$ is multivariate normal.  For the $f_{ip}$ we use a full multivariate normal evaluated at all times at which variable $p$ is observed for person $i$.  In order to evaluate the $f_{ip}$ at additional times e.g. during model evaluation we can use the conditional GP framework and treat the observed values as pseudo-inputs, from the sparse GP literature (\cite{Tit}).  Finally, for all model parameters $\Theta$ we learn a point estimate, so that their variational distributions are delta functions.  We impose vague normal priors on $ \Lambda^{(p)},\beta^{(p)},$ and $W$, a uniform prior on $\Psi^{(p)}$, and simply learn MLEs for other parameters.  Thus, the goal of our variational algorithm is to learn optimal variational parameters $\lambda_i = \{ \nu_{c_i}, \nu_{z_{ip}}, \mu_{b_i}, \Sigma_{b_i}, \mu_{f_{ip}}, \Sigma_{f_{ip}}\}$ for each individual $i$, as well as a point estimate $\hat{\Theta}$ for the model parameters.  In practice, we optimize the Cholesky decompositions of covariance matrices.

\subsubsection*{Solving the Optimization Problem}

We use the automatic differentiation package \textit{autograd} \footnote{https://github.com/HIPS/autograd} in Python to compute analytic gradients in order to optimize the lower bound, since the lower bound has an analytic expression.  At each iteration of the algorithm, we optimize the local variational parameters using exact gradients.  To optimize the global parameters, we turn to stochastic optimization, a commonly used tool in variational inference.  Rather than using the entire large dataset to compute exact gradients of the ELBO with respect to $\Theta$, we can compute an unbiased noisy gradient based on a sampled batch of observations (\cite{SVI}).  To set the learning rate we use RMSProp, which adaptively allows for a different learning rate for each parameter (\cite{rms}).

\section{Empirical Study}

In this section we describe our experimental setup and results on our dataset.

\subsubsection*{Dataset}

Our dataset contains laboratory values from 44,519 patients with stage 3 CKD or higher extracted from the Duke University Health System EHR.  IRB approval (\#Pro00066690) was obtained for this work.  We first created an initial cohort of roughly 600,000 patients that had at least one encounter in the health system in the year prior to Feb. 1, 2015.  This includes all types of encounters within the health system, including inpatient, outpatient, and emergency department visits, over a span of roughly 20 years.  From this, we filtered to patients who had at least five recorded values for serum creatinine, the laboratory value we used to calculate eGFR.  We next filtered to patients that had Stage 3 CKD or higher, indicative of moderate to severe kidney damage, defined as two eGFR measurements less than 60 mL/min separated by at least 90 days.  

We also choose to model five related lab values that have important clinical significance for CKD.  The first, Serum Albumin, is an overall marker of health and nutrition.  The second, Serum Bicarbonate, can indicate acid accumulation from inadequate acid elimination by the kidney.  The third, Serum Calcium, can indicate improperly functioning kidneys if levels are too high.  The fourth, Serum Phosphorus, can indicate phosphorus accumulation due to inadequate elimination by the kidney, and is associated with cardiovascular death and bone disorders.  Finally, Urine Albumin to Creatinine Ratio (ACR) is a risk factor and cause of kidney failure.  While all 44,519 patients have at least five eGFR measurements, there are 884, 242, 78, 16159, and 4321 patients who have no recorded values for the other five labs, respectively.  The median number of measurements for each lab (among patients with at least one) is 14, 8, 11, 12, 3, and 4, respectively.  


As a final preprocessing step, since the recorded eGFR values are extremely noisy and eGFR is only a valid estimate of kidney function at steady state, we take the mean of eGFR readings in monthly time bins for each patient.  Rapid fluctuations in acute illness are related to long term risk, but we have not yet explicitly incorporated this into our modeling.  We also do this to the other lab values, to reduce the overall noise in short time spans when there may be many rapidly changing measurements.  In order to align the patients on a common time axis, for each patient we fix $t=0$ to be their first recorded eGFR reading below 60 mL/min.  The baseline covariates used for $x_{i}$ were baseline age, race and gender, and indicator variables for hypertension and diabetes, as well as an overall intercept.

\subsubsection*{Evaluation}

After learning a point estimate for the global model parameters during training, they are held fixed.  Then, an approximate posterior over the local latent variables is learned for each patient in the held-out test set.  Predictions about future lab values are made by drawing samples from the approximate posterior predictive.  We compare our method with the method of (\cite{Schulam}), trained independently to each of the 6 labs.  We use 10 fold cross validation and use one-sided, paired t-tests to test for significant improvements in performance.  For each test patient, we learn their parameters using their data up until time $t=1$, $t=2$, and $t=4$, and record the mean absolute error of the predictions for each lab in future time bins.  These values are then averaged over all patients in the test set, and finally averaged over the 10 cross validation folds to produce Table 1.

\subsubsection*{Results}

\begin{table}[h]
  \caption{Mean Absolute Errors across all labs from 10 fold cross validation.  Bold indicates p-value from one-sided, paired t-test comparing methods was $<.05$.  *,**,*** indicate $p$ $<.01$, $<.001$, $<.0001$, respectively.}
  \label{sample-table}
  \centering
  \tiny 
  \begin{tabular}{| l l | l l l l | l l l | l l |}%
    \multicolumn{2}{c}{Predictions with data up to...} & \multicolumn{4}{c}{$t=1$} & \multicolumn{3}{c}{$t=2$} & \multicolumn{2}{c}{$t=4$}\\
    \hline
    \hline
    Lab & Model     & $(1,2]$  & $(2,4]$ & $(4,8]$ & $(8,19]$ & $(2,4]$ & $(4,8]$ & $(8,19]$ & $(4,8]$ & $(8,19]$ \\
    \hline
    eGFR & Schulam & \bf{8.86}*  &  \bf{10.43}* & \bf{12.05}  & \bf{13.69}  & \bf{8.84}*** & \bf{11.08}** & \bf{13.23}* & \bf{9.39}*** & \bf{12.29}*  \\
   & Proposed    & 9.12 &  10.67 & 12.28  & 14.21 & 9.26 & 11.73 & 13.99 & 10.12 & 13.07 \\ 
	\hline
    Serum Alb. & Schulam & 0.59 & 0.79 & 1.09 & 1.53 & 0.60 & 0.88 & 1.28 & 0.63 & 0.96 \\
    & Proposed & \bf{0.34}*** & \bf{0.39}*** & \bf{0.47}*** & \bf{0.63}*** & \bf{0.35}*** & \bf{0.45}*** & \bf{0.63}*** & \bf{0.40}*** & \bf{0.58}*** \\
    \hline
        Serum Bicarb. & Schulam & 1.92 & 2.06 & 2.13 & 2.31 & 1.93 & 2.06 & 2.21 & 1.89 & \bf{2.14}  \\
    & Proposed & 1.87 & 1.97 & 2.04 & 2.31 & 1.89 & 1.99 & 2.31 & 1.87 & 2.24 \\
    \hline
        Serum Calc.& Schulam & 0.74 & 1.02 & 1.62 & 2.89 & 0.72 & 1.26 & 2.27 & 0.85 & 1.53 \\
    & Proposed & \bf{0.37}*** & \bf{0.44}*** & \bf{0.58}*** & \bf{0.80}*** & \bf{0.39}*** & \bf{0.54}*** & \bf{0.80}*** & \bf{0.46}*** & \bf{0.73}***  \\
    \hline
        Serum Phos. & Schulam & 1.02 & 1.35 & 1.46 & 1.44 & 1.17 & 1.36 & 1.34 & 1.13 & 1.15  \\
    & Proposed & \bf{0.57}*** & \bf{0.68}*** & \bf{0.88}*** & \bf{1.23} & \bf{0.65}*** & \bf{0.88}*** & 1.25 & \bf{0.82}*** &1.23  \\
    \hline
        Urine ACR & Schulam & 1.17 & 1.30 & 1.44 & 1.64 & 1.14 & 1.30 & 1.53 & 1.11 & 1.41 \\
    & Proposed & \bf{0.92}*** & \bf{1.02}*** & \bf{1.17}*** & \bf{1.44} & \bf{0.96}*** & \bf{1.13}* & 1.45 & \bf{1.02} & 1.42  \\
    \hline
  \end{tabular}
\end{table}

Table 1 displays the results of both methods.  Our proposed method outperforms (\cite{Schulam}) for labs that had higher amounts of missingness.  In these settings, the methods sometimes have to predict held-out lab values for a patient without having yet observed any values for that lab.  In this case (\cite{Schulam}) can only predict using baseline covariates for that individual, while our multivariate approach is able to leverage information from the other related labs as well.  Interestingly, our method performs slightly worse at predicting eGFR, perhaps because this is the most commonly measured variable in our cohort.  Current clinical practice for managing care of CKD patients does not use any form of modeling to predict future disease status or lab values.  Incorporation of a model such as ours to predict future eGFR trajectory and other labs would provide a useful tool for nephrologists to assess the risk of future decline in kidney function in CKD patients.

\section{Discussion} 

\begin{figure*}
  \includegraphics[width=1.0\textwidth,height=19cm]{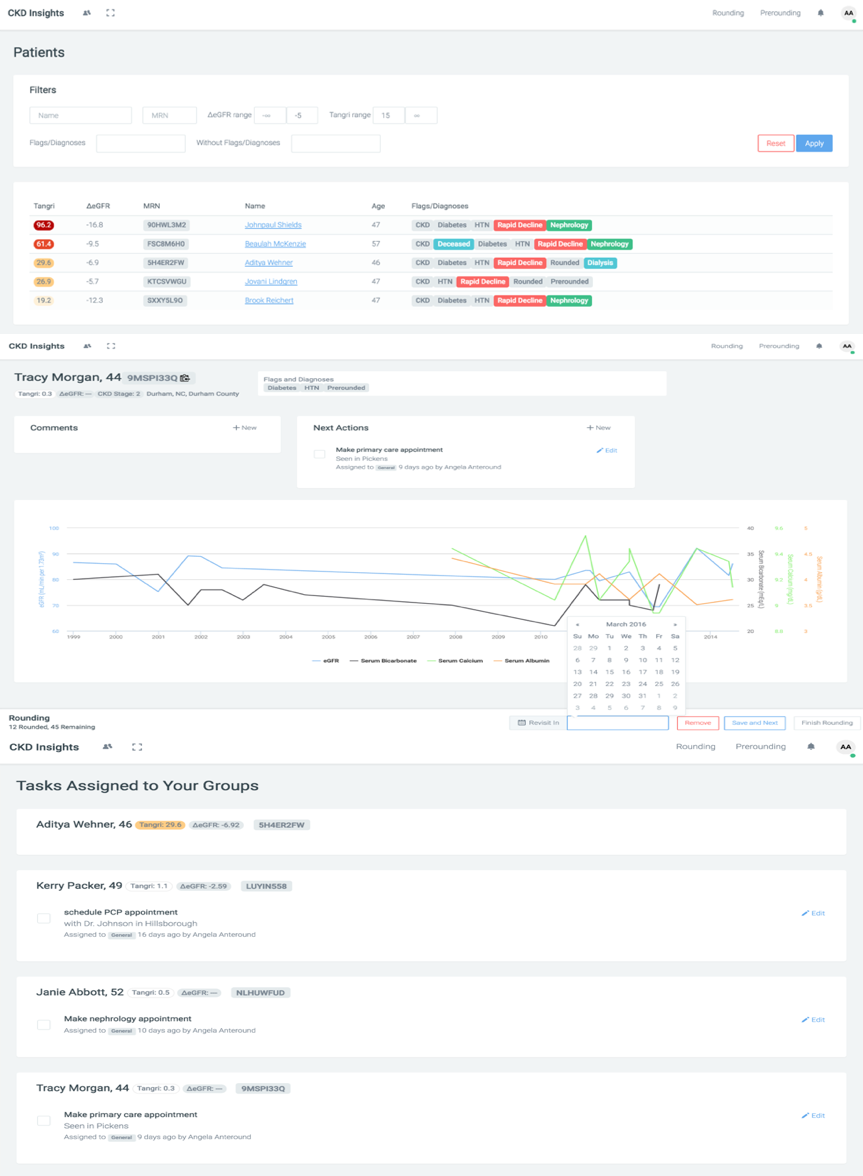}
  \caption{Snapshots from our CKD rounding application (with synthetic data).  The top panel shows a pre-rounding table of patients to be rounded on, along with risk scores and appropriate flags.  The middle panel displays patient data and other relevant information so that this patient's care can be efficiently managed and an appropriate intervention made, if applicable.  The bottom panel shows a list of tasks assigned to each group present at rounds.}
\end{figure*}

In this paper, we proposed a flexible model for multivariate longitudinal data, and applied it to disease trajectory modeling in CKD patients.  We find our model yields good performance on the task of predicting future kidney function and related lab values.  Although our work is a promising first step for developing machine learning models from EHR data and applying them to real clinical tasks, many inherent limitations to working with operational EHR data (\cite{Hersh}) must be overcome in order for the methods to be used in practice.

There are many interesting directions for extending the ideas presented here.  We plan to consider models for other diseases, including diabetes and cardiovascular disease, which are frequently comorbid with CKD.  Incorporation of a larger number of longitudinal variables will require care in order to ensure tractability, especially if we choose to model vitals that are recorded much more frequently.  Another interesting direction to consider is joint models for adverse event data, such as in (\cite{UAI}).  Jointly modeling multiple event processes (e.g. emergency department visits, heart attacks, strokes) will allow us to learn correlations between different types of events.  Given that much of the data recorded in the EHR is in the form of administrative billing codes, future work should incorporate these into the models as well.  Finally, models incorporating additional outcomes such as medical costs, hospitalizations, and patient quality of life are of significant practical interest.

The screenshot in Figure 2 shows a working prototype of the application being developed in collaboration with Duke Connected Care, where we use our predictions to help flag high risk patients for chart review during weekly rounding sessions.  By further refining and deploying a flexible, scalable predictive tool such as ours, ACOs around the country can intervene on high-risk patients and realize the potential benefits of precision medicine.  

 \acks{Joseph Futoma is supported by an NDSEG fellowship.  Dr. Cameron is supported by a Duke Training Grant in Nephrology (5T32DK007731).  This project was funded by both the Duke Translational Research Institute and the Duke Institute for Health Innovation.  Research reported in this publication was supported by the National Center for Advancing Translational Sciences of the NIH under Award Number UL1TR001117. The content is solely the responsibility of the authors and does not necessarily represent the official views of the NIH.}

\bibliography{mlhc}

\begin{thebibliography}{27}
\providecommand{\natexlab}[1]{#1}
\providecommand{\url}[1]{\texttt{#1}}
\expandafter\ifx\csname urlstyle\endcsname\relax
  \providecommand{\doi}[1]{doi: #1}\else
  \providecommand{\doi}{doi: \begingroup \urlstyle{rm}\Url}\fi

\bibitem[CMS(2016)]{CMS}
CMS.
\newblock Quality payment program: Delivery system reform, medicare payment
  reform, and macr, 2016.
\newblock URL
  \url{https://www.cms.gov/Medicare/Quality-Initiatives-Patient-Assessment-Instruments/Value-Based-Programs/MACRA-MIPS-and-APMs/MACRA-MIPS-and-APMs.html}.

\bibitem[Durichen and et~al.(2015)]{Durichen}
R.~Durichen and M.~Pimentel et~al.
\newblock Multitask gaussian processes for multivariate physiological
  time-series analysis.
\newblock In \emph{IEEE Transactions on Biomedical Engineering}, pages
  314--322, 2015.

\bibitem[Futoma et~al.(2016)Futoma, Sendak, Cameron, and Heller]{UAI}
J.~Futoma, M.~Sendak, C.~Cameron, and K.~Heller.
\newblock Scalable joint modeling of longitudinal and point process data for
  disease trajectory prediction and improving management of chronic kidney
  disease.
\newblock In \emph{UAI, to appear}, 2016.

\bibitem[Ghassemi and et~al.(2015)]{Ghassemi}
M.~Ghassemi and M.~Pimentel et~al.
\newblock A multivariate timeseries modeling approach to severity of illness
  assessment and forecasting in icu with sparse, heterogeneous clinical data.
\newblock In \emph{AAAI}, 2015.

\bibitem[Hersh and et~al(2013)]{Hersh}
W.~Hersh and M.~Weiner et~al.
\newblock Caveats for the use of operational electronic health record data in
  comparative effectiveness research.
\newblock In \emph{Medical Care}, 2013.

\bibitem[Hoffman et~al.(2013)Hoffman, Blei, Wang, and Paisley]{SVI}
M.~Hoffman, D.~Blei, C.~Wang, and J.~Paisley.
\newblock Stochastic variational inference.
\newblock In \emph{JMLR}, pages 1303--1347, 2013.

\bibitem[Johnson and et~al.(2015)]{Johnson}
T.~Johnson and D.Rinehart et~al.
\newblock For many patients who use large amounts of healthcare services, the
  need is intense yet temporary.
\newblock In \emph{Health Affairs}, 2015.

\bibitem[Lazarsfeld and Henry(1968)]{LSA}
P.~Lazarsfeld and N.~Henry.
\newblock Latent structure analysis.
\newblock In \emph{Houghton Mifflin}, 1968.

\bibitem[Levey and et~al.(2009)]{eGFR}
A.~Levey and L.~Stevens et~al.
\newblock A new equation to estimate glomerular filtration rate.
\newblock In \emph{Annals of Internal Medicine}, pages 604--612, 2009.

\bibitem[Lian and et~al.(2015)]{Lian}
W.~Lian and R.~Henao et~al.
\newblock A multitask point process predictive model.
\newblock In \emph{ICML}, 2015.

\bibitem[Mendelssohn et~al.(2011)Mendelssohn, Curtis, and et~al.]{Mend}
D.~Mendelssohn, B.~Curtis, and K.~Yeates et~al.
\newblock Suboptimal initiation of dialysis with and without early referral to
  a nephrologist.
\newblock In \emph{Nephrology Dialysis Transplantation}, 2011.

\bibitem[Murphy(2012)]{Murphy}
K.~Murphy.
\newblock Machine learning: A probabilistic perspective.
\newblock In \emph{MIT Press}, 2012.

\bibitem[Parikh et~al.(2016)Parikh, Kakad, and Bates]{Parikh}
R.~Parikh, M.~Kakad, and D.~Bates.
\newblock Integrating predictive analytics into high-value care: The dawn of
  precision delivery.
\newblock In \emph{JAMA}, pages 651--652, 2016.

\bibitem[Perotte and et~al.(2015)]{Perotte}
A.~Perotte and R.~Ranganath et~al.
\newblock Risk prediction for chronic kidney disease progression using
  heterogeneous electronic health record data and time series analysis.
\newblock In \emph{JAMIA}, 2015.

\bibitem[Proust-Lima and et~al.(2014)]{PL}
C.~Proust-Lima and et~al.
\newblock Joint latent class models for longitudinal and time-to-event data: A
  review.
\newblock In \emph{Statistical Methods in Medical Research}, 2014.

\bibitem[Ranganath and et~al.(2015)]{Raj}
R.~Ranganath and A.~Perotte et~al.
\newblock The survival filter: Joint survival analysis with a latent time
  series.
\newblock In \emph{UAI}, 2015.

\bibitem[Rizopoulos(2011)]{Riz}
D.~Rizopoulos.
\newblock Dynamic predictions and prospective accuracy in joint models for
  longitudinal and time-to-event data.
\newblock In \emph{Biometrics}, 2011.

\bibitem[Roberts and et~al.(2013)]{Shi}
S.~Roberts and M.~Osborne et~al.
\newblock Gaussian processes for time series modeling.
\newblock In \emph{Philosophical Transactions of the Royal Society A:
  Mathematical, Physical and Engineering Sciences}, 2013.

\bibitem[Saria and Goldenberg(2015)]{subtype}
S.~Saria and A.~Goldenberg.
\newblock Subtyping: What is it and its role in precision medicine.
\newblock In \emph{IEEE Intelligent Systems}, 2015.

\bibitem[Schulam and Saria(2015)]{Schulam}
P.~Schulam and S.~Saria.
\newblock A framework for individualizing predictions of disease trajectories
  by exploiting multi-resolution structure.
\newblock In \emph{NIPS}, 2015.

\bibitem[Shi et~al.(2005)Shi, Murray-Smith, and Titterington]{GPtime}
J.~Shi, R.~Murray-Smith, and D.~Titterington.
\newblock Hierarchical gaussian process mixtures for regression.
\newblock In \emph{Statistics and Computing}, 2005.

\bibitem[Smart et~al.(2014)Smart, Dieberg, Ladhanni, and Titus]{Smart}
N.~Smart, G.~Dieberg, M.~Ladhanni, and T.~Titus.
\newblock Early referral to specialist nephrology services for preventing the
  progression to end-stage kidney disease.
\newblock In \emph{Cochrane Database of Systematic Reviews}, 2014.

\bibitem[Szczech et~al.(2014)Szczech, Stewart, and et~al.]{Szczech}
L.~Szczech, R.~Stewart, and H.~Su et~al.
\newblock Primary care detection of chronic kidney disease in adults with
  type-2 diabetes: The add-ckd study.
\newblock In \emph{PLoS ONe}, 2014.

\bibitem[Tangri and et~al.(2011)]{Tangri}
N.~Tangri and L.~Stevens et~al.
\newblock A predictive model for progression of chronic kidney disease to
  kidney failure.
\newblock In \emph{JAMA}, pages 1553--1559, 2011.

\bibitem[Tieleman and Hinton(2012)]{rms}
T.~Tieleman and G.~Hinton.
\newblock Lecture 6.5-rmsprop: Divide the gradient by a running average of its
  recent magnitude, 2012.

\bibitem[Titsias(2009)]{Tit}
M.~Titsias.
\newblock Variational learning of inducing variables in sparse gaussian
  processes.
\newblock In \emph{AISTATS}, 2009.

\bibitem[Tuot et~al.(2011)Tuot, Plantinga, and et~al.]{Tuot}
D.~Tuot, L.~Plantinga, and C.~Hsu et~al.
\newblock Chronic kidney disease awareness among individuals with clinical
  markers of kidney dysfunction.
\newblock In \emph{Clinical Journal of the American Society of Nephrology},
  2011.

\end{thebibliography}


\end{document}